\documentclass[letterpaper, 10 pt, conference]{ieeeconf}  

\IEEEoverridecommandlockouts                              

\usepackage{color}

\newcommand{\policy}{{\pi}}
\newcommand{\expert}{{\pi_\beta}}

\newcommand{\clip}{\textrm{clip}}

\usepackage{graphics} 
\usepackage{times} 
\usepackage{amsmath} 
\usepackage{amssymb}  
\usepackage{amsfonts}

\usepackage{graphicx}
\usepackage[colorlinks=true,citecolor=black,linkcolor=black,urlcolor=blue]{hyperref}
\usepackage{fullpage}
\usepackage{subcaption}
\usepackage{caption}

\title{\LARGE \bf
Imitation Is Not Enough: 
Robustifying Imitation with Reinforcement Learning for Challenging Driving Scenarios

}

\author{Yiren Lu$^{1}$, Justin Fu$^{1}$, George Tucker$^{2}$, Xinlei Pan$^{1}$, Eli Bronstein$^{1}$, Rebecca Roelofs$^{2}$, Benjamin Sapp$^{1}$,\\ Brandyn White$^{1}$, Aleksandra Faust$^{2}$, Shimon Whiteson$^{1}$, Dragomir Anguelov$^{1}$, Sergey Levine$^{2,3}$
\noindent\thanks{$^{1}$ Waymo Research. %
$^{2}$ Google Research, Brain Team $^{3}$ UC Berkeley.
\newline Point of contact: {\tt\small maxlu@waymo.com}.
\newline Webpage: \href{https://waymo.com/research/imitation-is-not-enough-robustifying-imitation-with-reinforcement-learning/}{waymo.com/research/imitation-is-not-enough-robustifying-imitation-with-reinforcement-learning}
} 
}
\begin{document}

\maketitle
\thispagestyle{empty}
\pagestyle{empty}

\begin{abstract}

Imitation learning (IL) is a simple and powerful way to use high-quality human driving data, which can be collected at scale, to 
produce human-like behavior.
However, policies based on imitation learning alone often fail to sufficiently account for safety and reliability concerns.
In this paper, we show how imitation learning combined with reinforcement learning using simple rewards can substantially improve the safety and reliability of driving policies over those learned from imitation alone.
In particular, we train a policy on over 100k miles of urban driving data, and measure its effectiveness in test scenarios grouped by different levels of collision likelihood.
Our analysis shows that while imitation can perform well in low-difficulty scenarios that are well-covered by the demonstration data, our proposed approach significantly improves robustness on the most challenging scenarios (over 38\% reduction in failures). 
To our knowledge, this is the first application of a combined imitation and reinforcement learning approach in autonomous driving that utilizes large amounts of real-world human driving data.

\end{abstract}

\section{INTRODUCTION}

Building an autonomous driving system that is deployable at scale presents many difficulties. First and foremost is the challenge of handling the numerous rare and challenging edge cases that occur in real-world driving. To this end, imitative learning based approaches have been proposed that allow the performance of the method to scale with the amount of data available~\cite{pomerleau1988alvinn,bojarski2016end,codevilla2018end}. 
While situations that are well represented in the demonstration data are likely to be handled correctly by such a policy, more unusual or dangerous situations that occur only rarely in the data might cause the imitation policy -- which has not been explicitly instructed on what constitutes a risky or inappropriate response -- to respond unpredictably. 
The problem is compounded by complex interactions, where human expert driving data in similar scenarios may be scarce and sub-optimal~\cite{zhou2022long}.

Reinforcement Learning (RL) has the potential to resolve this by leveraging explicit reward functions that tell the policy what constitutes safe or unsafe outcomes (e.g., collisions).  Furthermore, because RL methods train in closed-loop, RL policies can establish causal relationships between observations, actions, and outcomes. This yields policies that are (1) less vulnerable to covariate shifts and spurious correlations commonly seen in open loop IL~\cite{ross2011reduction, de2019causal}, and (2) aware of safety considerations encoded in their reward function, but which are only implicit in the demonstrations.

\begin{figure}
  \begin{center}
    \includegraphics[width=0.41\textwidth]{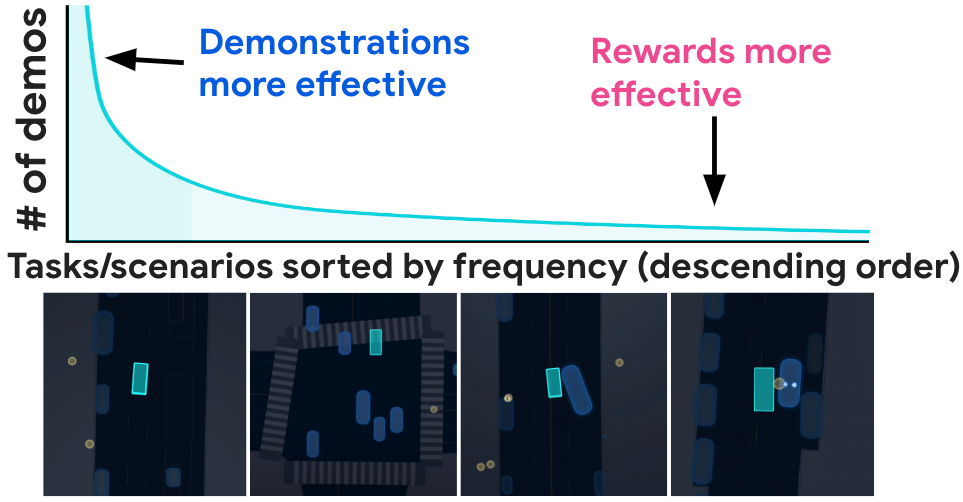}
  \end{center}
  \vspace{-0.15in}
  \caption{\small The demonstration-reward trade-off. As the amount of data for a particular scenario decreases, reward signals become more important for learning. We show a few visual examples representing scenarios with different frequencies.}
  \label{fig:demo_reward}
  \vspace{-0.3in}
\end{figure}

However, relying on RL alone, e.g., \cite{pan2017virtual,liang2018cirl,Zhang_2021_ICCV}, is also problematic because it heavily depends on reward design, which is an open challenge in autonomous driving~\cite{knox2021reward}. Without accounting for imitation fidelity, driving policies trained with RL may be technically safe but unnatural, and may have a hard time making forward progress
in situations that demand human-like driving behavior to coordinate with other agents and follow driving conventions.
IL and RL offer complementary strengths: IL increases realism and eases the reward design burden and RL improves safety and robustness, especially in rare and challenging scenarios in the absence of abundant data (Fig.~\ref{fig:demo_reward}). 

In this paper we focus on the driving scenarios that are most likely to exhibit safety and reliability concerns, leveraging the difficulty estimation from \cite{bronstein2022embedding}. Our proposed method, BC-SAC, combines IL and RL with a \emph{simple} reward function, and trains on difficult driving scenarios. Difficulty is estimated via a classifier that estimates the likelihood of a collision or near-miss when re-simulated with a pre-trained planning policy.  Our proposed reward function enforces safety of the agent, while natural driving behaviors are implicitly learned with IL.
The training data comes from a subset of real-world human driving data (over 100k miles of real-world urban driving data) \cite{bronstein2022embedding}. We demonstrate that this approach substantially improves the safety and reliability of policies learned over imitation alone without compromising on human-like behavior, showing 38\% and 40\% improvements over pure IL and RL baselines.

The main contributions of our work are: (1) We conduct the first large-scale application of a combined IL and RL approach in autonomous driving utilizing large amounts of real-world urban human driving data (over 100k miles) and a \emph{simple} reward function. (2) We systematically evaluate its performance and baseline performance by slicing the dataset by difficulty, demonstrating that combining IL and RL improves safety and reliability of policies over those learned from imitation alone (over 38\% reduction in safety events on the most difficult bucket).

\begin{table*}
\small
\centering
\caption{\small A comparison of different learning-based approaches to robotic control and autonomous driving.}
\begin{tabular}{|l||l|l|l|l|}
\hline
                          & Offline Demo      & Closed-loop & Rewards & Example Methods \\ \hline
         Behavior Cloning (BC) & Expert Demos      & No                   & No  & Multipath~\cite{chai2019multipath}, Precog~\cite{precog}, Trajectron++~\cite{salzmann2020trajectron++}   \\ \hline
Adversarial Imitation/IRL & Expert Demos      & Yes                  & No & IRL~\cite{ng2000algorithms}, GAIL~\cite{ho2016generative}, MGAIL~\cite{baram2016model}   \\ \hline
    RL                    & No               & Yes                  & Yes  & DQN~\cite{mnih2015human}, SAC~\cite{haarnoja18sac}  \\ \hline
              Offline RL  & Behavioral Data   & No                   & Yes  & CQL~\cite{kumar2020conservative}, TD3+BC~\cite{fujimoto2021minimalist}  \\ \hline
               ``Imitative'' RL  & Expert Demos      & Yes                  & Yes  & DQfD~\cite{hester2018deep}, DAPG~\cite{rajeswaran2018dapg}, BC-SAC (ours)  \\ \hline
\end{tabular}
\vspace{-0.2in}
\label{tab:prob_formulation}
\end{table*}

\section{RELATED WORK}



\textbf{Learning-based approaches in autonomous driving.}
We briefly summarize key properties of different learning-based algorithms for planning in Table~\ref{tab:prob_formulation}.
IL was among the earliest and most popular learning-based approaches adopted for deriving driving policies~\cite{pomerleau1988alvinn, bojarski2016end,zhang2021end,vitelli2022safetynet, nayakanti2022wayformer,lioutas2022titrated}. 
Controllable models trained with either IL~\cite{codevilla2018end,rhinehart2018deep} or RL~\cite{liang2018cirl} allow the user to specify high-level commands in the form of goals or control signals (e.g., left, right, straight) to combine higher-level route planning with low-level control. 

Two drawbacks of IL methods are: (1) open-loop IL (such as the widely used behavioral cloning approach~\cite{chai2019multipath,salzmann2020trajectron++,precog,liang2020laneGCN,ngiam21scene_transformer}) suffers from covariate shift~\cite{ross2011reduction} (which can be addressed with closed-loop training~\cite{ng2000algorithms,ho2016generative}), (2) IL methods lack \emph{explicit} knowledge of what constitutes good driving, such as collision avoidance.
RL methods have been proposed that allow the policy to learn from explicit reward signals with closed-loop training and have been applied to tasks such as lane-keeping~\cite{kendall2019learning}, intersection traversal~\cite{isele2018navigating}, and lane changing~\cite{wang2018reinforcement}. While these works show the efficacy of RL on specific scenarios, our work analyzes both the large-scale, aggregate performance \emph{and} challenging and safety-critical edge cases that make autonomous driving difficult to deploy in a real-world system. 

RL and other closed-loop methods for autonomous driving typically use simulation for training. There are a number of such public environments, which vary in how realistic they are, in particular what drives the simulated agents (e.g., expert-following/log playback~\cite{vinitsky2022nocturne,kothari2021drivergym,li2022metadrive,lioutas2022critic}, intelligent driving model (IDM) ~\cite{caesar2021nuplan}, or other rule based systems~\cite{dosovitskiy2017carla} and ML-based agents~\cite{caesar2021nuplan,ramamohanarao2016smarts}), and whether scenarios are procedurally generated (e.g., \cite{dosovitskiy2017carla,highway-env,ramamohanarao2016smarts}) or initialized from real-world driving scenes~\cite{zhan2019interaction, li2022metadrive, vinitsky2022nocturne,lioutas2022critic}. In our experiments, we develop and evaluate in closed-loop on real-world data with other agents following logs.


\textbf{Combining IL and RL.}
Methods such as DQfD~\cite{hester2018deep}, DDPGfD~\cite{vecerik2017leveraging}, and DAPG~\cite{rajeswaran2017learning} have shown that IL can help RL overcome exploration challenges in domains with known sparse rewards. Offline RL approaches, such as TD3+BC~\cite{fujimoto2021minimalist} and CQL~\cite{kumar2020conservative} combine RL objectives with IL ones to regularize $Q$-learning updates and avoid overestimating out-of-distribution values.
Our goal is not to propose a novel algorithmic combination of IL and RL, but rather to leverage this general approach to address challenges in autonomous driving at scale. 

\textbf{Addressing challenging and safety-critical scenarios for autonomous vehicles.}
~\cite{zhou2022long} learns policies that address long-tail scenarios in autonomous driving by using an ensemble of IL planners combined with model-predictive control. Another approach to improving safety is to augment a learned planner with a rule-based fallback layer that guarantees safety~\cite{shalev2016safe,vitelli2022safetynet}. Our work differs from these approaches, in that we directly incorporate safety awareness into the model learning process through a reward. Our method is also compatible with a fallback layer if needed, although we this is potential future work.
Another way to improve robustness of polices is to increase the frequency of negative examples during training. ~\cite{gandhi2017learning} collects failure data that covers various ways an unmanned aerial vehicle can crash, and the combined negative and positive data helps to train more robust policies. ~\cite{bronstein2022embedding} investigates the use of curriculum training to improve performance on challenging edge cases. While we also increase the exposure of the policy to challenging scenarios during training, we extend these findings by showing how RL yields outsized improvements on the hardest scenarios.


\section{BACKGROUND}

\subsection{Markov Decision Processes (MDPs)}

In this work, we cast the autonomous driving policies learning problem as a Markov decision process. Following standard formalism, we define an MDP as a tuple $\{\mathcal{S}, \mathcal{A}, \mathcal{T}, \mathcal{R}, \gamma, \rho_0\}$. $\mathcal{S}$ and $\mathcal{A}$ denote the state and action spaces, respectively. $\mathcal{T}$ denotes to transition model. $\mathcal{R}$ represents the reward function, and $\gamma$ represents the discount factor. $\rho_0$ represents the initial state distribution. The objective is to find a policy $\pi$, a (stochastic) mapping from $\mathcal{S}$ to $\mathcal{A}$, that maximizes the expected discounted sum of rewards,
$ \pi^* = \max_\pi \mathbb{E}_{\mathcal{T}, \pi, \rho_0} \left[ \sum_{t=0}^\infty \gamma^t R(s_t, a_t) \right].  $

\subsection{Imitation Learning (IL)}
IL constructs an optimal policy by mimicking an expert. We assume an expert (an optimal policy), denoted as $\expert$, produces a dataset of trajectories $\mathcal{D} = \{s_0, a_0, \cdots, s_N, a_N\}$ through interaction with the environment. The learner's goal is to train a policy $\pi$ that imitates the $\expert$. In practice, we only observe the expert states, so we estimate expert actions using inverse dynamics. For example, behavioral cloning (BC) trains the policy via a log-likelihood objective,
$ \mathbb{E}_{s, a \sim \mathcal{D}}\left[ \log \pi(a|s) \right]$.
Alternatively, closed loop approaches include inverse RL (IRL)~\cite{ng2000algorithms} and adversarial IL (GAIL~\cite{ho2016generative}, MGAIL~\cite{baram2016model}), which instead aim to more directly match the occupancy measure or state-action visitation distribution between the policy and the expert, rather than indirectly through the conditional action distribution. In principle, this can resolve the covariate shift issue that affects open loop imitation~\cite{ross2011reduction}.

\subsection{Reinforcement Learning (RL)}
\label{sec:background_rl}
RL aims to learn an optimal policy through an iterative, online trial and error process. In this work we use off-policy, value-based RL algorithms such as $Q$-learning. These methods aim to learn the state-action value function, defined as the expected future return when starting from a particular state and action:
\[{\displaystyle Q^\pi(s, a) =  \mathbb{E}_{\mathcal{T}, \pi, \rho_0} \left[ \sum_{t=0}^\infty \gamma^t R(s_t, a_t) | s_0 = s, a_0 = a \right]}.  \]
%
%
In this work, we use an actor-critic method for training continuous control policies. Typical actor-critic methods alternate between training a critic $Q$ to minimize the Bellman error and an actor $\pi$ to maximize the value function. We use the entropy-regularized updates of Soft Actor-Critic (SAC)~\cite{haarnoja18sac}:
\begin{align}
& \min_{Q}\ \mathbb{E}_{s, a , s'\sim \pi} [(Q(s, a) - \hat{Q}(s, a, s')  )^2] 
\label{eq:sac_critic_loss}
\end{align}
\begin{equation} 
\max_{\pi}\ \mathbb{E}_{s, a \sim \pi}\left[Q(s, a) + \mathcal{H}(\pi(\cdot | s))\right], 
\label{eq:sac_actor_loss}
\end{equation}
\noindent where
\begin{align}
\hat{Q}(s, a, s') = r(s, a) + \gamma \mathbb{E}_{a' \sim {\pi}}\left[ \bar{Q}(s', a') - \log \pi(a'|s')\right]
\end{align}
\label{eq:sac_critic_target}
\noindent and $\bar{Q}$ denotes a target network that is a copy of the critic through which gradients do not pass.

\section{Learning to Drive with RL-Augmented BC}


We wish to design an approach that benefits from the complementary strengths of IL and RL. Imitation provides an abundant source of learning signal without the need for reward design, and RL addresses the weaknesses of IL in rare and challenging scenarios where data is scarce. Following this intuition, we formulate an objective that utilizes the learning signal from demonstrations where data is abundant and the reward signal where data is scarce. Specifically, we utilize a weighted mixture of the IL and RL objectives:
\begin{equation}
\max_\pi\ \mathbb{E}_{\mathcal{T}, \pi, \rho_0} \left[ \sum_{t=0}^\infty \gamma^t R(s_t, a_t) \right]  
+ \lambda \mathbb{E}_{s, a \sim \mathcal{D}}[
\log \pi (a|s)].
\label{eq:weighted_objective}
\end{equation}
\subsection{Behavior Cloned Soft Actor-Critic (BC-SAC)}
\begin{figure}
  \begin{center}
    \includegraphics[width=0.4\textwidth]{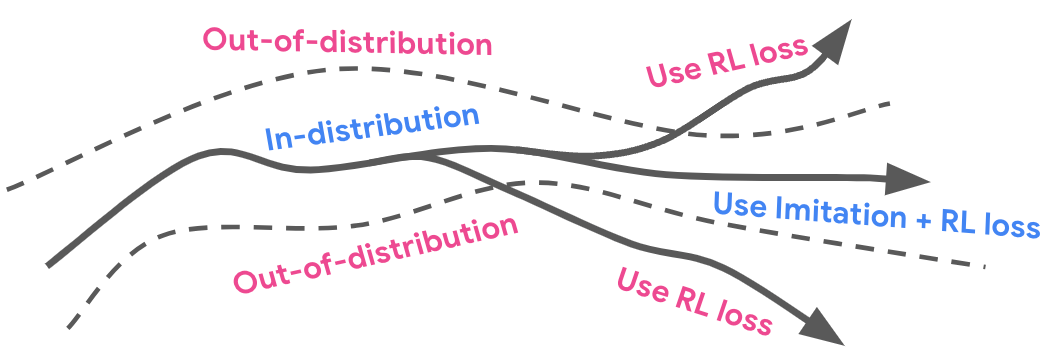}
  \end{center}
  \caption{\small Different objective influence. For in-distribution states, both IL and RL objectives provide learning signal. For out-of-distribution states, the RL objective dominates.}
  \label{fig:method_overview}
  \vspace{-0.25in}
\end{figure}

While in principle a variety of RL methods could be combined with IL to optimize Eq.~\ref{eq:weighted_objective}, a convenient choice for efficient training is to use actor-critic algorithms, in which case the policy can be optimized with respect to Eq.~\ref{eq:weighted_objective} simply by adding the imitation learning objective to the expected value of the Q-function (i.e., the critic), similarly to DAPG~\cite{rajeswaran2018dapg} or TD3+BC~\cite{fujimoto2021minimalist}. Building on the widely used SAC framework, which further adds an entropy regularization objective to the actor, we obtain our full actor objective:
\begin{align*}
\begin{split}
  \mathbb{E}_{s, a \sim \pi}[Q(s, a) + \mathcal{H}(\pi(\cdot | s))] + \lambda \mathbb{E}_{s, a \sim \mathcal{D}}[\log\policy(a|s)].
\end{split}
\end{align*}
The critic update remains the same as in SAC, outlined in Eq~\ref{eq:sac_critic_loss}. With the appropriate setting of $\lambda$, 
this objective encourages the policy to mimic the expert data when it is within the data distribution $\mathcal{D}$. However, in out-of-distribution states the policy primarily relies on reward to learn. Fig.~\ref{fig:method_overview} visualizes this concept.

\subsection{Reward Function}
While designing a reward function to capture ``good'' driving behavior is an open-challenge~\cite{knox2021reward}, we can side-step this issue by relying on the imitation learning loss to primarily guide the policy, while the simple reward function only needs to encode safety constraints.
To this end, we use a combination of collision and off-road distances as our reward signal.
The collision reward is 
\begin{align}
R_\text{collision} = \min(d_{\text{collision}}-d_{\text{c\_offset}}, 0),
\label{eq:collsion_rew}
\end{align}
where $d_{\text{collision}}$ is the Euclidean distance in meters of the closest points between the ego vehicle and a nearest bounding box of other vehicles; $d_{\text{c\_offset}}$ (default 1.0) is an offset added to encourage the vehicle to keep a distance from nearby objects. The off-road reward is 
\begin{align}
R_{\text{off-road}}=\clip(-d_{\text{o\_offset}}-d_{\text{to-edge}}, -2.0, 0.0),
\label{eq:offroad_rew}
\end{align}
where $d_{\text{to-edge}}$ is the distance in meters of the vehicle to the nearest road edge (negative being on-road, positive being off-road). $d_{\text{o\_offset}}$ (default 1.0) is an offset to encourage the vehicle to keep a distance to road edge. We combine the rewards additively, such that
$
R=R_{\text{collision}}+R_{\text{off-road}}.$

\subsection{Forward and Inverse Vehicle Dynamics Models}
\label{sec:dynamics}

We update the vehicle's state using the kinematic bicycle dynamics model ~\cite{rajamani2011vehicle}, which computes the vehicle's next pose $(x, y, \theta)$ given a steering and acceleration action $a = (a_\textrm{steer}, a_\textrm{accel})$. In order to obtain expert actions for imitation learning, we use an inverse dynamics model to solve for the actions that would have achieved the same states as the logged trajectories in our dataset. These expert actions are found by minimizing the MSE of the corners' $(x, y)$ positions between the inferred state $\mathcal{T}(s_t, a_t)$ and ground-truth next state $s_{t+1}$.

\subsection{Model Architecture}
We use a dual actor-critic architecture similar to TD3 and SAC~\cite{fujimoto2018addressing,haarnoja18sac}: the main components are an actor network $\pi(a|s)$, a double $Q$-critic network $Q(s, a)$ and a target double $Q$-critic network $\bar{Q}(s, a)$. Each network has a separate Transformer observation encoder described in~\cite{jaegle2021perceiver}
that encodes features including all vehicle states, road-graph points, traffic lights signals, and route goals. The actor network outputs a $\tanh$-squashed diagonal Gaussian distribution parameterized by a mean $\mu$ and variance $\sigma$. 

\subsection{Training on Difficult Examples}

The performance of learning-based methods strongly depends on the training data distribution,
especially in safety-critical settings with long-tail distributions~\cite{frank2008reinforcement,karla2016, shalev2016safe,aloq,FPO}).
Autonomous driving falls in this category: most scenarios are mundane, but a sizable minority of scenarios have critical safety concerns. Following \cite{bronstein2022embedding}, which demonstrated that training on more difficult examples results in better performance than using unbiased training distributions, we explore how the training distribution affects the method performance. 



\section{EXPERIMENTS}
\label{sec:experiments}


\subsection{Experimental Setup}

\noindent\textbf{Datasets}. We use a dataset (denoted \textbf{All}) consisting of over 100k miles of expert driving trajectories, split into 10 second segments, collected from a fleet of vehicles operating in San Francisco (SF) \cite{bronstein2022embedding}.
We divide these segments into 6.4 million for training and 10k for testing.
Trajectories from the same vehicle operating on the same day are stored in the same partition to avoid train-test leakage.
The trajectories, which are sampled at 15 Hz, contain features describing the autonomous vehicle (AV) state and the state of the environment
as measured by the AV's perception system.
We use the \emph{difficulty model} described by \cite{bronstein2022embedding} as a proxy for measuring the rarity of events, since it is difficult to directly construct a scenario-level out-of-distribution estimator, and challenging scenarios are generally less frequent. 
Given a run segment, the difficulty model predicts whether a segment will result in a collision or near-miss when re-simulated with an internal AV planner.
We trained the difficulty model in a supervised manner using cross-entropy loss on a dataset consisting of 5.6k positive examples and 80k negative examples, with binary human labels.
We create the \textbf{Top1}, \textbf{Top10}, and \textbf{Top50} subsets by selecting the top 1\% (40k train, 1.2k test), 10\% (400k train, 19k test), and 50\% (2 million train, 66k test) percentiles of difficulty model scores from a chronologically separate dataset of 4 million segments, respectively.

\noindent\textbf{Simulation}. As mentioned in Sec.~\ref{sec:dynamics}, vehicle dynamics are modeled using a 2D bicycle dynamics model. The behavior of other vehicles and pedestrians in the scene are replayed from the logs (log-playback), similarly to~\cite{vinitsky2022nocturne,kothari2021drivergym,li2022metadrive}. While this means that agents are non-reactive, it ensures that the behavior of other agents is human-like, and the inclusion of imitative losses discourages the learned policy to deviate too far from the logs, which would cause the log-playback agents to become unrealistic. We also use short segments of 10s to mitigate pose divergence. 


\noindent\textbf{Baselines}. We compare our method to both open-loop (\emph{BC}~\cite{pomerleau1988alvinn}) and closed-loop
 (\emph{MGAIL}~\cite{baram2016model}) imitative methods. The latter takes advantage of closed loop training and the differentiability of the simulator dynamics. For completeness, we also include a SAC baseline to represent an RL-only approach. 

\noindent\textbf{Metrics}.
We evaluate agents using two metrics:
\begin{enumerate}
\item \emph{Failure Rate}: Percentage of the run segments that have at least one \emph{Collision} or \emph{Off-road} event at any timestep. \emph{Collision} is true if the bounding box of the ego vehicle intersects with a bounding box of another object. \emph{Off-road} is true if the bounding box of the ego vehicle deviates from the drivable surface according to the map.
\item \emph{Route Progress Ratio}: Ratio of the distance traveled along the route by the policy compared to the expert demonstration. We project the ego vehicle's state onto the route and compute the total length from the start of the route.
\end{enumerate}

\begin{figure}
  \begin{center}
    \includegraphics[width=0.45\textwidth]{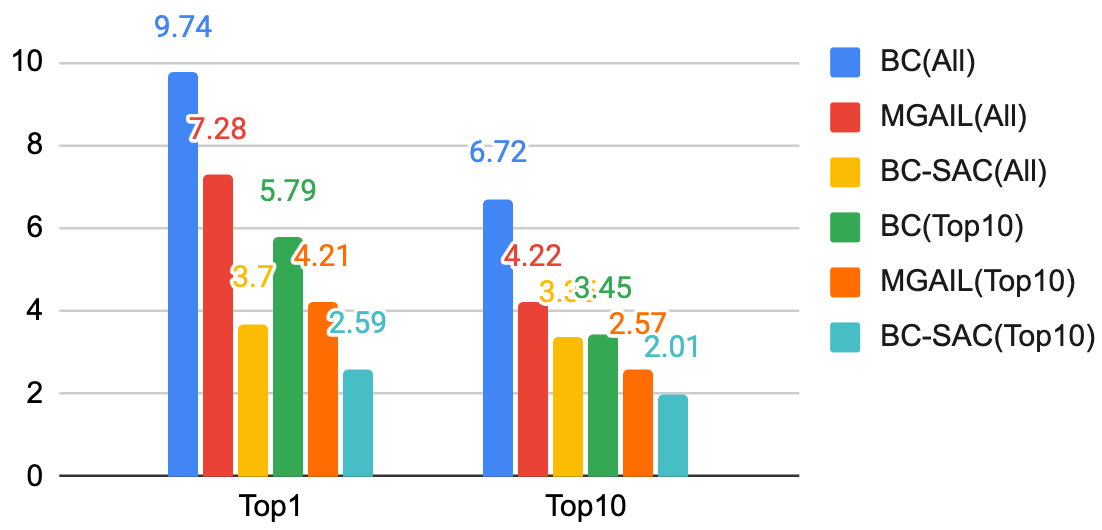}
  \end{center}
  \vspace{-0.1in}
  \caption{\small Failure rates on the most challenging evaluation sets: Top1 and Top10 (lower is better, with training on All and Top10). BC-SAC consistently achieves the lowest error rates.}
  \label{fig:top1_top10}
  \vspace{-0.2in}
\end{figure}

\begin{figure}
  \begin{center}
    \includegraphics[width=0.5\textwidth]{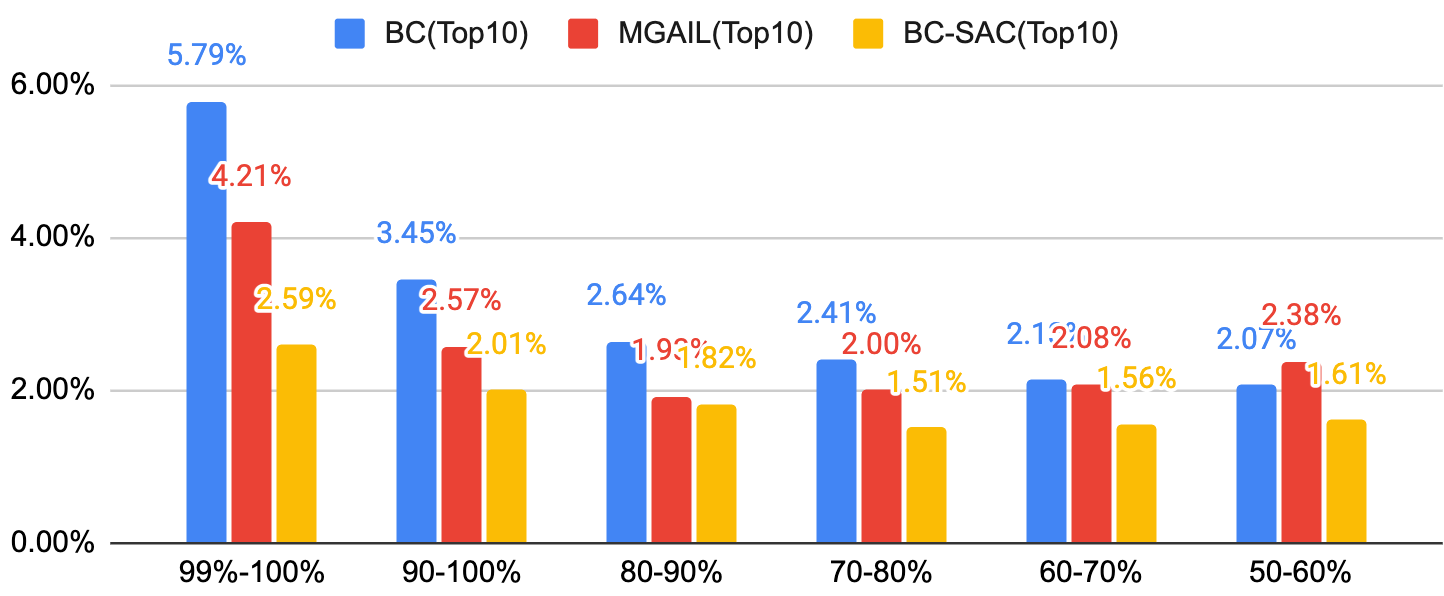}
  \end{center}
  \vspace{-0.1in}
  \caption{
  \small Failure rates of BC, MGAIL, and BC-SAC across scenarios of varying difficulty levels (50\%-100\%, lower is better). 
  While all methods perform worse as the evaluation dataset becomes more challenging, BC-SAC always performs best and shows the least degradation.}
  \label{fig:all_risky_buckets}
  \vspace{-0.2in}
\end{figure}

\begin{figure*}
\centering
\begin{minipage}{.7\textwidth}
  \centering
  \footnotesize
\begin{tabular}{c|c|ccccccc|cc}
\hline
         Method
         & Training
         & \begin{tabular}[c]{@{}c@{}}Top1 (\%)\end{tabular} 
         & \begin{tabular}[c]{@{}c@{}}Top10 (\%)\end{tabular}  
         & \begin{tabular}[c]{@{}c@{}}Top50 (\%)\end{tabular} 
         & \begin{tabular}[c]{@{}c@{}}All (\%)\end{tabular}
         & \begin{tabular}[c]{@{}c@{}}Route Progress \\ Ratio, All(\%)\end{tabular} \\
\hline

BC & All 
&{9.74\scriptsize$\pm$0.49}
&{6.72\scriptsize$\pm$0.47}
&{5.14\scriptsize$\pm$0.39}
&{4.35\scriptsize$\pm$0.27}
&{99.00\scriptsize$\pm$0.39}
\\

MGAIL & All 
&{7.28\scriptsize$\pm$0.98}
&{4.22\scriptsize$\pm$0.77}
&{3.40\scriptsize$\pm$0.97}
&\bf{2.48\scriptsize$\pm$0.29}
&\bf{99.55\scriptsize$\pm$1.91}
\\

SAC & All 
&{5.29\scriptsize$\pm$0.66}
&{4.64\scriptsize$\pm$1.08}
&{4.12\scriptsize$\pm$0.74}
&{6.66\scriptsize$\pm$0.44}
&{77.82\scriptsize$\pm$8.21}
\\

BC-SAC & All 

&\bf{3.72\scriptsize$\pm$0.62}
&\bf{2.88\scriptsize$\pm$0.23}
&\bf{2.64\scriptsize$\pm$0.21}
&{3.35\scriptsize$\pm$0.31}
& {95.26\scriptsize$\pm$8.64}
\\
\hline

BC & Top10 
&{5.79\scriptsize$\pm$0.82}
&{3.45\scriptsize$\pm$0.72}
&{2.71\scriptsize$\pm$0.57}
&{3.64\scriptsize$\pm$0.31}
&\bf{98.06\scriptsize$\pm$0.18}
\\

MGAIL & Top10 
&{4.21\scriptsize$\pm$0.95}
&{2.57\scriptsize$\pm$0.52}
&{2.20\scriptsize$\pm$0.52}
&\bf{2.45\scriptsize$\pm$0.35}
&{96.57\scriptsize$\pm$1.19}
\\

SAC & Top10 
&{4.33\scriptsize$\pm$0.47}
&{4.11\scriptsize$\pm$0.63}
&{3.66\scriptsize$\pm$0.47}
&{5.60\scriptsize$\pm$0.86}
&{71.05\scriptsize$\pm$2.47}
\\

BC-SAC & Top10 

&\bf{2.59\scriptsize$\pm$0.31}
&\bf{2.01\scriptsize$\pm$0.29}
&\bf{1.76\scriptsize$\pm$0.20}
&{2.81\scriptsize$\pm$0.26}
&{87.63\scriptsize$\pm$0.58}
\\

\hline

BC & Top1 
&{7.66\scriptsize$\pm$1.13}
&{7.84\scriptsize$\pm$0.92}
&{6.63\scriptsize$\pm$0.78}
&{6.85\scriptsize$\pm$0.65}
&\bf{94.10\scriptsize$\pm$1.00}
\\

MGAIL & Top1 
& {4.24\scriptsize $\pm$0.95} 
& {3.16\scriptsize $\pm$0.43}
& {2.74\scriptsize $\pm$0.46}
& {3.79\scriptsize $\pm$0.46} 
& {93.10\scriptsize $\pm$11.72}\\

SAC & Top1
&{4.15\scriptsize$\pm$0.31}
&{3.87\scriptsize$\pm$0.12}
&{3.46\scriptsize$\pm$0.16}
&{5.98\scriptsize$\pm$1.03}
&{75.63\scriptsize$\pm$2.19} \\

BC-SAC & Top1 
&\bf{3.61\scriptsize$\pm$0.87}
&\bf{2.96\scriptsize$\pm$1.11}
&\bf{2.69\scriptsize$\pm$0.87}
&\bf{3.38\scriptsize$\pm$0.48}
&{75.00\scriptsize$\pm$17.21}\\

\hline
\end{tabular}
\captionof{table}{\small Failure rates (lower is better) and progress ratios (higher is better) \\of BC-SAC and baselines on different training/evaluation subsets.}
  \label{table:main_results}
\end{minipage}%
\begin{minipage}{.3\textwidth}
  \centering
  \begin{subfigure}{0.45\linewidth}
    \centering
    \captionsetup{justification=centering}
    \includegraphics[width=\textwidth]{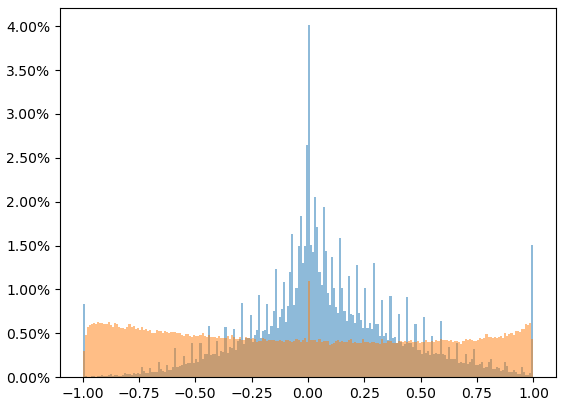}
    \caption{SAC\\ acceleration}
\end{subfigure}
\begin{subfigure}{0.45\linewidth}
    \centering
    \captionsetup{justification=centering}
    \includegraphics[width=\textwidth]{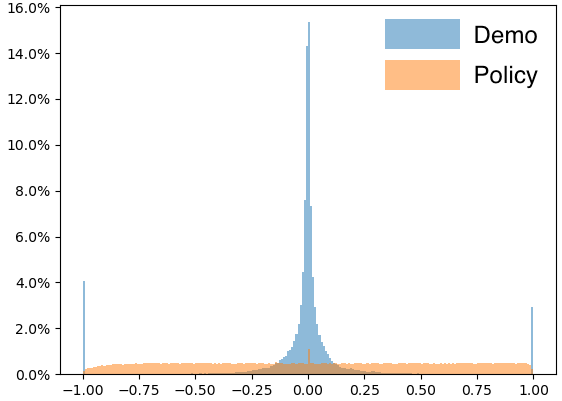}
    \caption{SAC tire\\ angles}
\end{subfigure}
\begin{subfigure}{0.45\linewidth}
    \centering
    \captionsetup{justification=centering}
    \includegraphics[width=\textwidth]{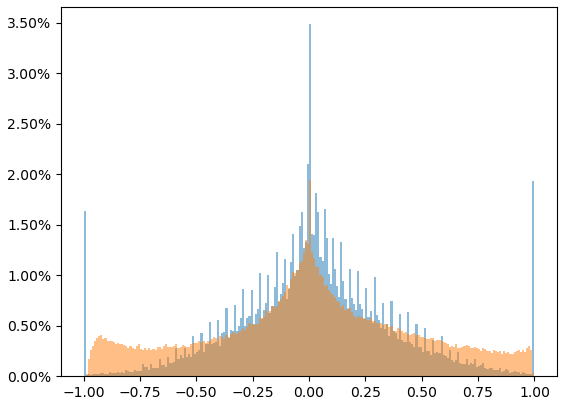}
    \caption{BC-SAC\\ acceleration}
\end{subfigure}
\begin{subfigure}{0.45\linewidth}
    \centering
    \captionsetup{justification=centering}
    \includegraphics[width=\textwidth]{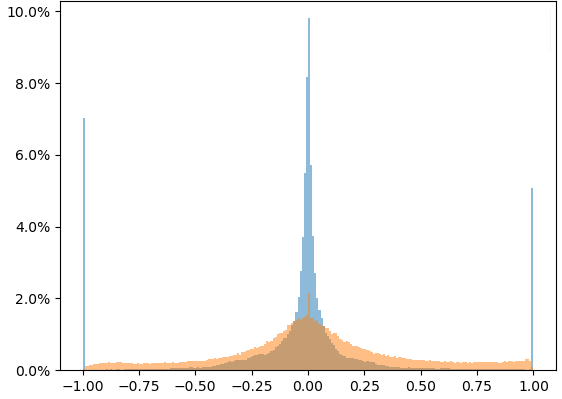}
    \caption{BC-SAC tire\\ angles}
\end{subfigure}
  \captionof{figure}{\small Marginal action distributions. SAC/BC-SAC (orange) vs logs (blue).
  }
  \label{fig:action_dist}
\end{minipage}
\vspace{-0.2in}
\end{figure*}

\subsection{Results}
We evaluate the baseline methods (BC, MGAIL, SAC) and our method (BC-SAC) trained on several subsets of the training dataset (All, Top10, and Top1), and evaluate against subsets of the evaluation set (Top1, Top10, Top50, All) in Table~\ref{table:main_results}. All configurations are evaluated with three random seeds, reporting mean and standard deviation. Previously, \cite{bronstein2022embedding} showed that training MGAIL on Top10 yields similar performance with training on All. Similarly, we find that all methods perform best when trained on Top10. Notably, BC trained on Top1 performs significantly worse compared to training on All or Top10, which reflects the fact that imitation learning methods rely on large amounts of data to implicitly infer driving preferences. 
In contrast, BC-SAC performs robustly when trained on Top1. 
Given that all methods perform best when trained on Top10, we focus on that setting in the following subsections. 

\noindent\textbf{BC-SAC comparison to imitation methods (BC, MGAIL) in the challenging scenarios.}
Figure~\ref{fig:all_risky_buckets} compares BC-SAC against BC and MGAIL across the evaluation dataset slices according to difficulty levels.
BC-SAC achieves better performance overall, especially in the more challenging slices where the performance of both BC and MGAIL substantially degrade. 
Additionally, BC-SAC has the lowest variance across scenarios of varying difficulty in performance ($\sigma=0.37$) vs. BC ($\sigma=1.29$) and MGAIL ($\sigma=0.78$). 

\begin{figure}
\begin{subfigure}{0.23\textwidth}
  \begin{center}
    \includegraphics[width=0.9\textwidth]{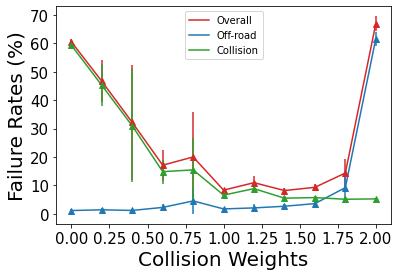}
  \end{center}
\end{subfigure}
\begin{subfigure}{0.23\textwidth}
  \begin{center}
    \includegraphics[width=0.9\textwidth]{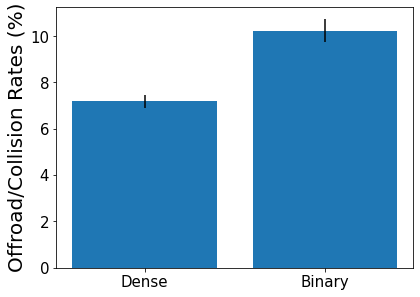}
  \end{center}
\end{subfigure}
  \caption{\small \textbf{Left}: Off-road / collision weights. Off-road weight and collision weight add up to 2.0. The x-axis is the collision weight. A balanced choice of off-road and collision weights lead to the best performance. 
  \textbf{Right}: Dense vs binary rewards. Binary reward is defined as $-1$ when a safety event happens and $0$ otherwise. Dense rewards lead to fewer safety events.
  }
  \label{fig:dense_discrete_rew}
  \vspace{-0.2in}
\end{figure}

\begin{figure}
\begin{subfigure}{0.23\textwidth}
      \begin{center}
        \includegraphics[width=0.9\textwidth]{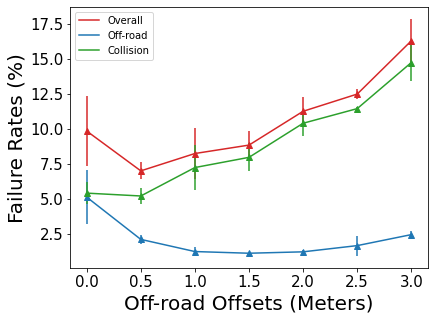}
      \end{center}
\end{subfigure}
\begin{subfigure}{0.23\textwidth}
  \begin{center}
    \includegraphics[width=0.98\textwidth]{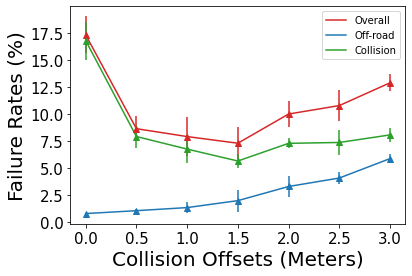}
  \end{center}
\end{subfigure}
  \caption{
\small  Off-road offset $d_{\text{o\_offset}}$ and collision offset $d_{\text{c\_offset}}$ ablations. A small amount of offsets improves overall performance.
}
  \vspace{-0.25in}
  \label{fig:offroad_collision_offsets}
\end{figure}

\begin{figure}
\begin{subfigure}{0.23\textwidth}
  \begin{center}
    \includegraphics[width=0.8\textwidth]{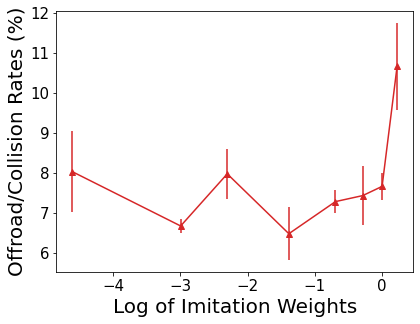}
  \end{center}
\end{subfigure}
\begin{subfigure}{0.23\textwidth}
  \begin{center}
    \includegraphics[width=0.9\textwidth]{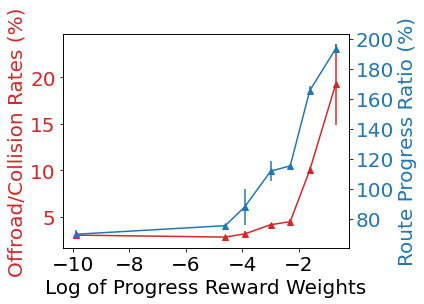}
  \end{center}
\end{subfigure}
  \caption{\small \textbf{Left} Imitation weights (log-scale) vs failure rates.
  \textbf{Right} Progress reward weights (log-scale) vs policy evaluation performance: safety event rate and route progress ratio.}
  \label{fig:imitation_lambda}
  \vspace{-0.2in}
\end{figure}

\noindent\textbf{BC-SAC comparison to RL-only training (SAC)}.
In all configurations, BC-SAC outperforms SAC in terms of safety metrics (Table~\ref{table:main_results}), likely because BC-SAC also utilizes learning signal from large amount of demonstrations.
SAC generates actions that deviate significantly from the demonstrations with more boundary action values yielding unnatural (more swerves) and uncomfortable (abrupt acceleration) driving behavior (Figure \ref{fig:action_dist}). With a BC loss, BC-SAC generates an action distribution similar to the logs. 
\noindent\textbf{Reward shaping and RL / IL weights}. We conduct a set of ablation studies to answer  how the form of the reward function and the weights on the RL and imitation components influence final performance. We use a smaller dataset constructed by sampling 10\% of the Top10 data and compare: (1) our full reward vs. a discrete binary reward (Fig~\ref{fig:dense_discrete_rew} Right), (2) off-road and collision reward term weights (Fig~\ref{fig:dense_discrete_rew} Left), (3) off-road and collision offset parameters (Fig~\ref{fig:offroad_collision_offsets}), and (4) the weight on the RL and IL terms in the objective (Fig~\ref{fig:imitation_lambda}). The results indicate that the proposed shaped reward improves overall performance over the simpler sparse reward with an appropriate choice of reward parameters, and a balance between imitation and RL terms leads to the best performance.

\noindent\textbf{Progress-safety balance}. While our work focuses on safety-critical scenarios, in Fig~\ref{fig:imitation_lambda} Right, we show that introducing a small amount of a progress reward leads to significantly more progress without major regressions in safety metrics. However, large progress rewards lead to degradation in performance.

\begin{figure}[ht]
  \centering
    \includegraphics[width=0.42\textwidth]{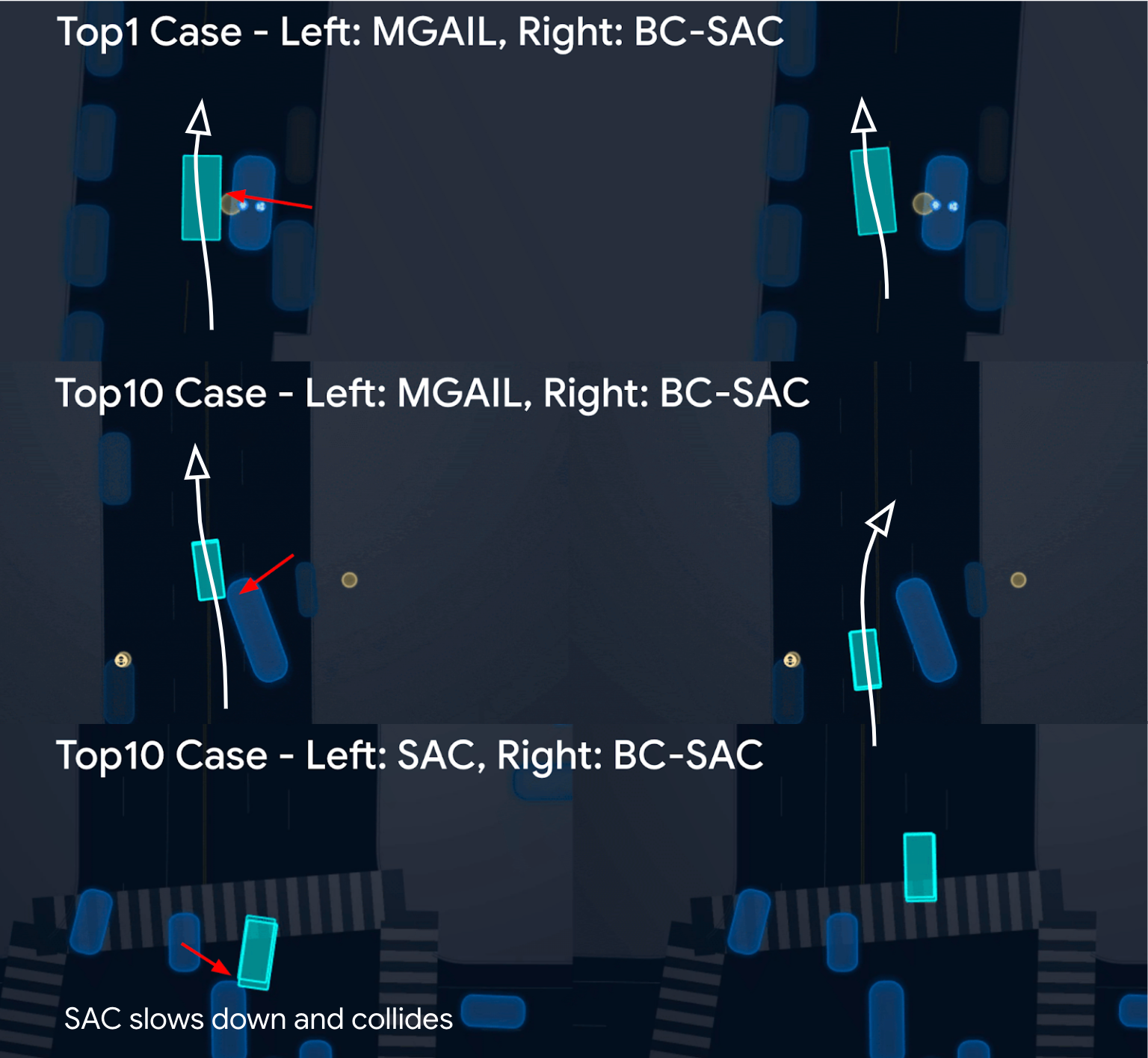}
  \vspace{-0.05in}
  \caption{
  \small Visualizations of a few scenarios where BC-SAC improves over imitation (MGAIL) and RL-only (SAC). The cyan car is controlled.  \textbf{Example 1}: MGAIL collides with a pedestrian exiting a double parked car while BC-SAC leaves enough clearance. \textbf{Example 2}: MGAIL does not provide sufficient clearance and collides with the incoming vehicle. \textbf{Example 3}: SAC slows down in an intersection resulting in an rear collision. BC-SAC maintains a proper speed profile through the intersection without a collision.
  }
  \label{fig:win_case1}
  \vspace{-0.2in}
\end{figure}

\begin{table}
\small
\centering
\begin{tabular}{|l||l|l|l|l|l|l|}
\hline
Method & CLIP & COLL & OFF & RED & LAN & DIV  \\ \hline
BC-SAC & 8 & 7 & 2 & 1 & 7 & 15 \\ \hline
MGAIL & 16 & 8 & 8 & 2 & 0 & 6 \\ \hline

\end{tabular}
\vspace{-0.05in}
\caption{\small \label{tab:qual_results}Failure frequency categorizations, per type, incurred by BC-SAC and MGAIL on a small sample set (N=80). BC-SAC generally has fewer direct collisions and off-road events, but has a greater frequency of being hit by other objects.}
\vspace{-0.25in}
\end{table}

\noindent\textbf{In-depth failure analysis.} Table~\ref{tab:qual_results} presents a detailed analysis of failure modes on a set of 80 sampled scenarios from the Top1 and Top10 buckets. We categorize failures into 6 broad buckets. CLIP (clipping): small collisions that occur when a vehicle collides with an object on the side while moving. OFF (off-road): failures when the agent drives off the road. LAN (bad lane): an agent encroaches into another lane, either the wrong lane or a bad merge, which results in a collision. COLL (collision): collision where the planning agent is at fault and drives into another vehicle. RED (red light): red light violations that result in collisions. Finally, DIV (log divergence): collisions where a sim agent collides with the planning agent due to divergence from the logs.

Overall, MGAIL tends to have more clipping collisions and off-road events. Fig~\ref{fig:win_case1} shows two of the cases where RL improves over IL. We hypothesize that our method improves in these cases because MGAIL, as an imitation method, lacks an explicit penalty for collisions, and thus is not sensitive to small collisions during otherwise realistic behavior. On the other hand, the collisions encountered by BC-SAC tend to be cases where the collision is not directly the result of the AV planner's action, but the planner diverges from the logs in a way such that it is hit by other vehicles. Because BC-SAC also is not explicitly rewarded for following traffic rules (though it inherits this behavior via imitation), we also see a small amount of failures due to that.

\section{CONCLUSIONS}
We presented a method for robust autonomous driving in challenging driving scenarios, that combines imitation learning with RL (BC-SAC), paired with a simple safety reward, and trained on large datasets of real-world driving. Overall, the method significantly improves safety and reliability in challenging scenarios, resulting in more than 38\% reduction in safety events of the most difficult scenarios compared to IL-only and RL-only baselines. Our extensive experiments examined the roles of training datasets, reward shaping and IL / RL objective terms. BC-SAC inherits implicit human-like driving behaviors from imitation, while RL is a fail-safe for handling out-of-distribution safety scenarios. 
Similarly to the IL-only settings, training on the top 10\% of the most challenging scenarios 
yields the most robust performance in the combined IL and RL setting. 
While this work mainly focused on optimizing safety-related rewards, a natural extension is to incorporate other factors into the objective, such as progress, traffic rule adherence, and passenger comfort. 
Besides the reward function, this approach does not account for unexpected behavior of other agents in response to out-of-distribution actions on the part of the ego vehicle, and it still requires heuristically choosing the tradeoff between the IL and RL objectives. A promising future work direction would be to enable reactive sim agents for training and evaluation and to extend the approach to enforce safety as an explicit constraint,
perhaps in combination with methodology to mitigate distributional shift.
\bibliographystyle{IEEEtran}
\bibliography{references}


\clearpage

\onecolumn

\appendix
\section{Appendix}
\subsection{IL + RL Distributed Actor-Learner Training Architecture}
\label{sec:actor_learner_arch}

\begin{figure*}[!ht]
  \centering
    \includegraphics[width=0.8\textwidth]{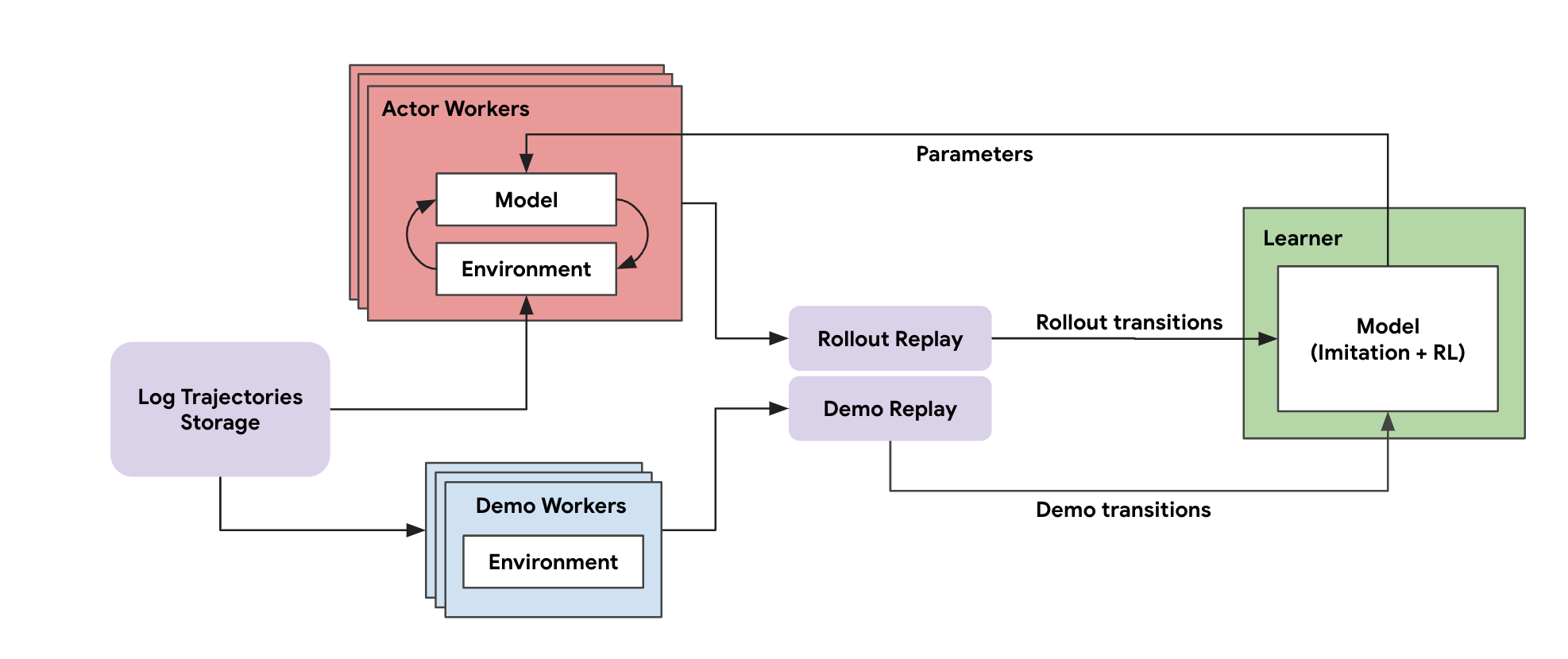}
  \caption{IL + RL distributed actor-learner training architecture. We extend the distributed IMPALA architecture
  ~\cite{espeholt2018impala} 
  with additional demo rollout workers and a demo replay buffer, which produce rollout transitions in the same format as the actor workers. The learner worker samples from both the rollout replay buffer and the demo replay buffer to perform training updates in an off-policy manner.}
  \label{fig:actor_learner}
\end{figure*}

\subsection{Additional Details on Model Architectures and Hyper-parameters Settings}
\label{sec:bc_sac_hparams}
We use a dual actor-critic architecture similar to TD3 and SAC~\cite{fujimoto2018addressing,haarnoja18sac}: each of the main components, actor network $\pi(a|s)$, double $Q$-critic network $Q(s, a)$ and target double $Q$-critic network $\bar{Q}(s, a)$, has a separate Transformer observation encoder described in~\cite{bronstein2022hierarchical}, and the encoder embedding is fed to a $(256, 256)$ fully connected head.
The actor network outputs a $\tanh$-squashed diagonal Gaussian distribution parameterized by a mean $\mu$ and variance $\sigma$. 

We train the BC-SAC algorithm with the following hyper-parameters: the actor learning rate is 1e-4, the critic learning rate is 1e-4, the imitation learning rate is 5e-5, the batch size is 64, and the reward discount ratio is 0.92. The sample-to-insert ratio for replay is 8, which is the average number of times the learner should sample each item in the replay buffer during the item's entire lifetime. In practice, instead of performing a combined gradient step of both the IL and RL objectives, we alternate the training steps between IL and RL with different update frequencies. For every 8 RL updates, we update with IL loss for one time. The hyper-parameters are found by performing grid-search.

For SAC, we use the same network design and hyper-parameters as in BC-SAC, except that it does not perform IL step.

For BC, we discretize the 2d action space (steer, acceleration) in to $31\times7 = 217$ actions with the same underlying dynamics model. 
We use a similar network design for BC as in BC-SAC's actor network with a Softmax prediction head representing probabilities of the discrete actions.
We use the cross-entropy loss with a learning rate of 1e-4 and batch size of 256 for training.

For MGAIL, we follow the network design and hyper-parameters setting presented in~\cite{bronstein2022hierarchical}.

\end{document}